\documentclass{article}

\usepackage[numbers]{natbib}
\usepackage[preprint]{neurips_2024}

\usepackage[utf8]{inputenc} % allow utf-8 input
\usepackage[T1]{fontenc}    % use 8-bit T1 fonts
\usepackage{hyperref}       % hyperlinks
\usepackage{url}            % simple URL typesetting
\usepackage{booktabs}       % professional-quality tables
\usepackage{amsfonts}       % blackboard math symbols
\usepackage{nicefrac}       % compact symbols for 1/2, etc.
\usepackage{microtype}      % microtypography
\usepackage{xcolor}         % colors
\usepackage{amsmath}
\usepackage{graphicx}
\usepackage{enumitem}
\usepackage{caption}
\usepackage{subcaption}
\usepackage{algorithm}
\usepackage{algpseudocode}
\usepackage{wrapfig} 
\usepackage{multirow}
\usepackage{arydshln}
\usepackage{pifont}
\usepackage[normalem]{ulem}
\usepackage{kotex}
\usepackage{listings}
\usepackage{color}
\usepackage{makecell}

\useunder{\uline}{\ul}{}

\definecolor{codegreen}{rgb}{0,0.6,0}
\definecolor{codegray}{rgb}{0.5,0.5,0.5}
\definecolor{codepurple}{rgb}{0.58,0,0.82}
\definecolor{backcolour}{rgb}{0.95,0.95,0.92}

\lstdefinestyle{mystyle}{
    backgroundcolor=\color{backcolour},   
    commentstyle=\color{codegreen},
    keywordstyle=\color{magenta},
    numberstyle=\tiny\color{codegray},
    stringstyle=\color{codepurple},
    basicstyle=\ttfamily\footnotesize,
    breakatwhitespace=false,         
    breaklines=true,                 
    captionpos=b,                    
    keepspaces=true,                 
    numbers=left,                    
    numbersep=5pt,                  
    showspaces=false,                
    showstringspaces=false,
    showtabs=false,                  
    tabsize=2
}

\title{HLQ: Fast and Efficient Backpropagation\\ via Hadamard Low-rank Quantization}

\author{%
  Seonggon Kim\quad Eunhyeok Park\\
  Department of Computer Science and Engineering, POSTECH\\
  \texttt{\{sungonuni, eh.park\}@postech.ac.kr} \\
}

\begin{document}

\maketitle

\begin{abstract}
With the rapid increase in model size and the growing importance of various fine-tuning applications, lightweight training has become crucial. Since the backward pass is twice as expensive as the forward pass, optimizing backpropagation is particularly important. However, modifications to this process can lead to suboptimal convergence, so training optimization should minimize perturbations, which is a highly challenging task. In this study, we introduce a novel optimization strategy called Hadamard Low-rank Quantization (HLQ), focusing on reducing the cost of backpropagation in convolutional and linear layers. We first analyze the sensitivity of gradient computation with respect to activation and weight, and judiciously design the HLQ pipeline to apply 4-bit Hadamard quantization to the activation gradient and Hadamard low-rank approximation to the weight gradient. This combination was found to be the best for maximizing benefits, and our extensive experiments demonstrate the outstanding performance of HLQ in both training from scratch and fine-tuning, achieving significant memory savings and acceleration on real GPUs with negligible quality degradation.
\end{abstract}

\section{Introduction}

Up until now, most efforts in model optimization have focused on reducing inference costs, with less attention given to the costs of training the model, since these costs are incurred only once. While it is crucial to optimize the inference process used repeatedly, training optimization has been relatively neglected because it must meet more challenging conditions. In inference optimization, one can easily estimate the quality loss induced by optimization and mitigate it through additional training, such as quantization-aware training (QAT)\cite{choi2018pact, bhalgat2020lsq+, park2020profit} or knowledge distillation\cite{chen2021distilling, wu2022causal, huang2022knowledge}. In contrast, performance degradation in training optimization is difficult to gauge and hard to recover from. Training optimization must reduce costs significantly while minimizing perturbations during training.

Due to the aforementioned limitations, training optimization has not been extensively explored. However, as models grow larger and the need for continuous updates across diverse applications increases\cite{wang2024comprehensive,hu2021lora}, the demand for efficient training has rapidly risen. The emergence of large-scale models, such as large language models (LLMs)\cite{touvron2023llama, radford2018improving} and Vision Transformers (ViTs)\cite{dosovitskiy2020image, li2022efficientformer, liu2021swin} accelerates this trend, making efficient training in resource-constrained environments 
imperative.

In response to this need, we explore new potential for training optimization. Unlike some prior approaches~\cite{xi2023training, chmiel2022accurate}, we focus exclusively on the backpropagation (BP) of linear and convolution layers. The rationale behind this strategy is twofold: first, the backward pass of these layers is twice as computationally expensive as the forward pass, accounting for the majority of the BP overhead; second, utilizing the unmodified forward pass guarantees precise loss evaluation, ensuring stable and optimal convergence. By designing an efficient BP pipeline with minimal perturbation, we can maintain the quality of the trained model while significantly reducing training overhead. 

In this study, we propose a novel efficient training scheme called Hadamard Low-rank Quantization (HLQ). This scheme selectively applies Hadamard Quantization (HQ)~\cite{xi2023training} and Hadamard Low-rank Approximation (HLA)~\cite{yang2024efficient}, taking into account the sensitivity of gradients in both activation and weight, to maximize the benefits. Overall, our contributions can be summarized into the following four items: 
\begin{itemize}
    \item Our observation shows that HLA is suitable for weight gradient optimization, but the activation gradient path suffers heavily with HLA.
    \item We also validate that the activation gradient is highly robust to low-precision HQ, but the weight gradient is much more vulnerable to HQ.
    \item Based on this observation, we design a novel idea called HLQ that applies HQ and HLA selectively to ensure stable and efficient training.
    \item We implement HLQ on a customized CUDA kernel and realize the benefits on real GPUs.   
\end{itemize}
According to our extensive analysis, HLQ sustains the accuracy of the trained model on complex datasets while offering benefits during training, achieving up to 2.5 times faster performance and an 78.59\% lower memory footprint.

\section{Related Work}
\subsection{Quantization for Training Optimization}
Quantization is one of the most successful optimization methods, and many studies have explored its potential for efficient training. The representative approach applies quantization to the forward pass. In \cite{chmiel2022accurate}, the authors aim to address the long tail of gradients by introducing a logarithmic quantizer with a custom FP4 data format. While they succeed in achieving high accuracy, their method is difficult to accelerate due to the customized representation. In \cite{xi2023training}, the authors apply HQ for the forward pass and structured pruning to the gradient for efficient training, showing promising results on language tasks. Additionally, alternative studies such as FP8 training \cite{micikevicius2022fp8, peng2023fp8lm} and integer-only training \cite{wang2020niti} have been actively studied, aiming to replace all numerical representations for both the forward and backward passes jointly. Unlike these previous studies, we focus solely on optimizing the backward pass while utilizing the vanilla forward pass to maintain the quality of training.

\subsection{Low-rank Approximation for Training Optimization}

In addition to quantization, there are other approaches focused on reducing the cost of training by decreasing the size of tensors. One prominent example is LoRA \cite{hu2021lora}, a lightweight fine-tuning technique for LLMs. LoRA adds low-rank decomposed weights alongside the pre-trained original weights and trains only the added weights, significantly reducing the training cost. There have also been studies aiming to achieve additional efficiency by applying quantization to the pre-trained weights \cite{dettmers2024qlora,xu2023qa}. However, these studies primarily focus on inference efficiency, with limited cost reduction during training. LBP-WHT \cite{yang2024efficient} was the first to optimize the training process by incorporating HLA into the backward pass. By reducing the rank along the sequence length or batch dimension in the path of computing activation and weight gradients, it significantly reduces the cost of the backward pass, which is twice as expensive as the forward pass. While it is more accurate and delivers higher training performance than LoRA, it still suffers from considerable accuracy loss compared to full-rank training, limiting its applicability. Our analysis revealed that activation gradients are highly sensitive to low-rank approximation, and based on this finding, we improved the optimization pipeline to achieve much higher efficiency while enhancing the quality of the trained network.

\section{Preliminary}
Before introducing HLQ, we explain the notation we use and provide relevant background knowledge.

\subsection{Backpropagation of Linear Layer}
Since we are dealing with multidimensional tensors, we will use unified dimension notation to aid understanding. We denote \(I\) for input channel, \(O\) for output channel, \(B\) for batch dimension, \(W\) for spatial width, \(H\) for spatial height, and \(L\) for sequence length. For brevity, we primarily explain our idea based on optimizing the backward pass for a linear layer, which consumes the majority of computation and memory space in transformer-based structures. However, please note that a convolutional layer can be optimized using the same technique by considering the spatial dimensions as the sequence length or replacing \(L\) with \(H \cdot W\).

In the forward pass, when an input \(x \in \mathbb{R}^{B \times L \times I}\) and a weight \(w \in \mathbb{R}^{O \times I}\) are given, the output \(y \in \mathbb{R}^{B \times L \times O}\) is produced by matrix multiplication \(y = x \cdot w^T\). In the backward pass, when the gradient of the output \(g_y \in \mathbb{R}^{B \times L \times O}\) is propagated from the following layer, we can calculate the gradients of the activation, \(g_x \in \mathbb{R}^{B \times L \times I}\), and weight, \(g_w \in \mathbb{R}^{O \times I}\), via the chain rule as follows:
\begin{equation}
g_w = \frac{1}{B} \bar{g}_y^T \cdot \bar{x}, \quad g_x = g_y \cdot w,
\label{equ:bwd}
\end{equation}
where \(\bar{g}_y \in \mathbb{R}^{(B \times L) \times O}\) and \(\bar{x} \in \mathbb{R}^{(B \times L) \times I}\) are the reshaped tensors of \(g_y\) and \(x\), respectively. Since each pass requires the same computation as the forward operation, it is desirable to optimize both.

\subsection{Hadamard Transform}
The Hadamard matrix\cite{sylvester1867lx} is an orthogonal matrix where the matrix is square, and its rows and columns are orthonormal vectors. Additionally, the elements of the Hadamard matrix are either \(+1\) or \(-1\), simplifying matrix multiplication through addition or subtraction operations.

The Hadamard transform (HT) is a linear mapping function that employs a special Hadamard matrix, known as the Walsh-Hadamard matrix, which can be viewed as a generalized version of the Fourier transform. The Walsh-Hadamard matrix \(H_d\) for the \(2^d\) dimension is defined as follows:
\begin{equation}
    \mathrm{H}_1 = \frac{1}{\sqrt{2}} \begin{bmatrix}
 1 & 1 \\ 
 1 & -1 
\end{bmatrix}, \quad \mathrm{H}_{n} = \mathrm{H}_1 \otimes \mathrm{H}_{n-1},
\end{equation}
where \(\otimes\) denotes the Kronecker product. When a vector \(\mathbf{x} \in \mathbb{R}^{2^d}\) is given, it can be mapped to the frequency domain via HT with \(H_d\), which can be performed with only \(\mathcal{O}(n \cdot \log n)\) addition/subtraction operations using the fast Walsh-Hadamard Transformation (FWHT)\cite{shanks1969computation}, where $n=2^d$.

In practice, a block-diagonal transformation, also known as the order-\(n\) 2D HT, is often used \cite{xi2023training}. Here, a block-diagonal matrix is \(\mathrm{H}_D = \text{BlockDiag}(H_k, \ldots, H_k)\), where \(D\) is a multiple of \(2^k\). The block-diagonal transformation can be seen as a local transformation, where the \(D\)-dimensional side of the data is reshaped into \(2^k \times \lceil D/2^k \rceil\), and independent HT are applied to the \(2^k\) side. This is preferred due to the efficiency on real GPUs, and we also utilize this as well.

Recently, HT has garnered significant attention for its ability to compensate for the quality loss induced by optimization techniques with minimal additional computational cost, and this benefit is applicable to both quantization and low-rank approximation. In the subsequent sections, we present an in-depth analysis of the strengths and weaknesses of HQ and HLA in the backward pass, followed by a detailed explanation of our approach based on these observations.

\section{Analysis on Hadamard Low-rank Approximation for Backpropagation}

\begin{figure}
    \centering
    \includegraphics[width=\textwidth]{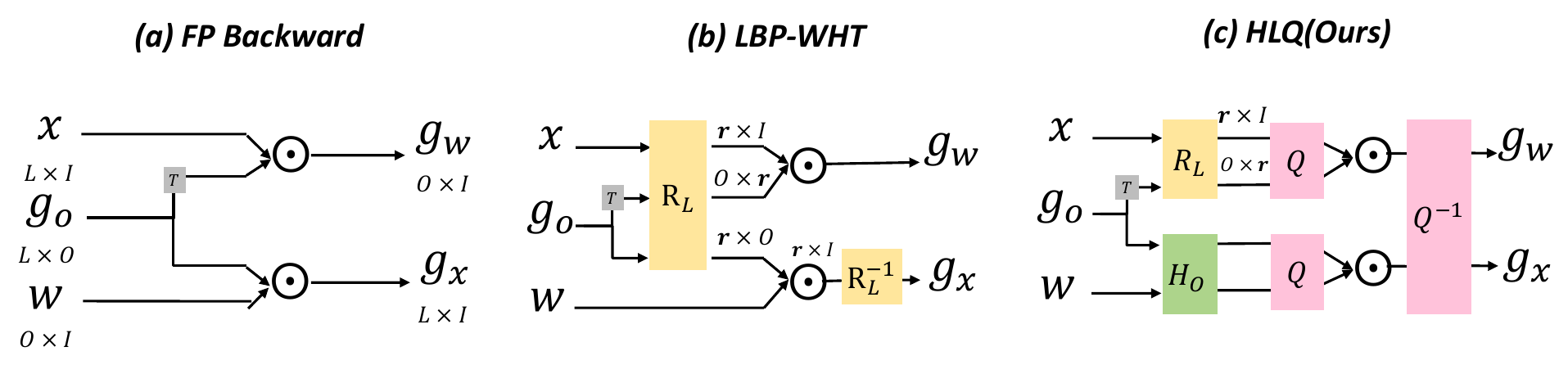}
    \vspace{-1.5\baselineskip}
    \caption{Overview of the backward pass for (a) vanilla pass, (b) LBP-WHT, and (c) HLQ.}
    \label{fig:HLQ}
    \vspace{-\baselineskip}
\end{figure}

Low-rank BackPropagation via Walsh-Hadamard Transformation (LBP-WHT)\cite{yang2024efficient} is a state-of-the-art study that applies HLA to backpropagation, aiming to replace LoRA\cite{hu2021lora} for parameter-efficient fine-tuning of ViT models by reducing update costs. This study demonstrates the potential of HLA, but it also reveals limitations: accuracy decreases somewhat during fine-tuning and significantly during training from random initialization. In this section, we extensively analyze LBP-WHT to present the limitations of HLA for backpropagation.

Figure \ref{fig:HLQ} (b) visualizes the LBP-WHT pipeline. This pipeline projects the $L$ dimension of the output gradient $g_y$ and the input activation $x$ into a lower rank $r << L$ using HLA. When the spatial resolution is small, the batch dimension $B$ is projected instead. Let the reduced output gradient and input activation be denoted as $\hat{g}_y\in \mathbb{R}^{r\times O}$ and $\hat{x}\in \mathbb{R}^{r\times I}$, respectively. The corresponding approximated gradient for the weight $\hat{g}_w\in \mathbb{R}^{O\times I}$ and input $\hat{g}_x \in \mathbb{R}^{r\times I}$ can be calculated as follows:
\begin{equation}
    \hat{g}_w = \hat{g}_y^t \cdot \hat{x}, \quad \hat{g}_x = \hat{g}_y^t \cdot w.
\end{equation}
Since $\hat{g}_w$ has the same dimensions as the target weight $w$, the gradient can be seamlessly integrated into the weight update. However, since $\hat{g}_x$ has a mismatched shape to $x$, the inverse projection function $R_L^{-1}$ needs to be applied to recover the original dimension size.

They observe that, because HT is a generalized Fourier transform, selecting an appropriate lower number of transform bases allows the carefully selected frequency components to create a representative low-rank representation. This property is helpful to sustain fine-tuning quality with significantly lowered BP cost, where the original computation requires $4OIL$ flops, while the modified approach takes only $(O + I)Lr + 4OIr + ILr$ flops.

\subsection{Limitation of LBP-WHT}\label{sec:LBP_limit}
According to their observations, LBP-WHT outperforms LoRA in terms of both computation cost and fine-tuning quality in ViT. However, their fine-tuning quality is inferior to full-rank training, and they are not applicable for training from scratch due to large accuracy degradation.

To exemplify the source of their limitations in detail, we conduct additional experiments by selectively applying LBP-WHT to \(g_x\) and \(g_w\) while training ResNet-50 on CIFAR-100 from scratch. In our setting, the model with the conventional BP shows a top-1 accuracy of 76.46\%. When we apply HLA to \(g_w\), the model shows 76.29 \% accuracy, which is only 0.17 \% lower than the original BP. However, when we apply HLA to \(g_x\), the top-1 drops to 72.01 \%, indicating a 4.45 \% accuracy degradation. 

Unlike \(g_w\), the \(g_x\) path is highly vulnerable to low-rank approximation. It appears that spatial (and batch-wise) discrepant information is crucial for maintaining the training quality of \(g_x\). However, LBP-WHT on \(g_x\) neglects spatial resolution via low-rank projection, and this information loss propagates through preceding layers during BP, resulting in notable accuracy degradation. 
In contrast, \(g_w\) is naturally averaged across the spatial and batch dimensions during updates. Therefore, the rank reduction in these dimensions seems to cause negligible damage. %Furthermore, conventional methods like momentum or adaptive learning rates smooth out the jitter in the gradient trajectory during stochastic descent, so selecting low-frequency components of \(g_w\) doesn't cause significant harm during training. 
To maximize the quality of updates with LBP-WHT, it is crucial to avoid applying low-rank approximation to \(g_x\).

\begin{figure}
    \centering
    \includegraphics[width=\textwidth]{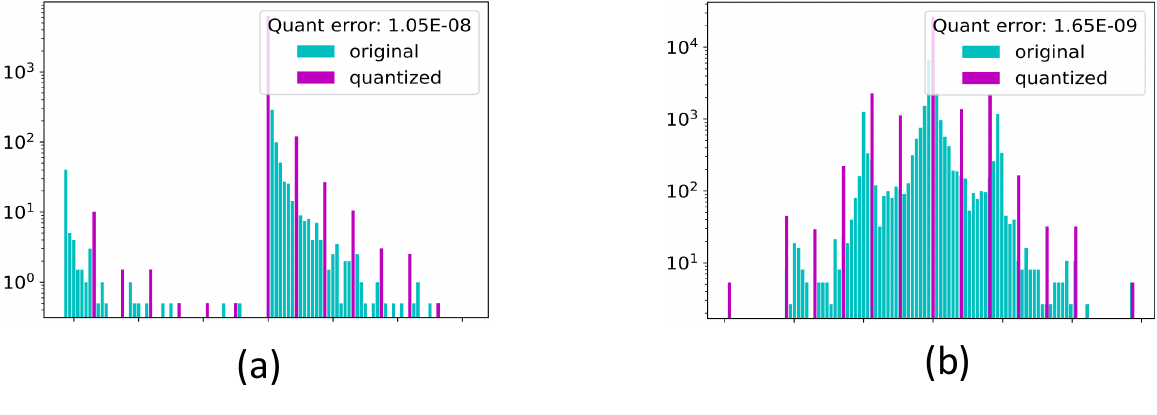}
    \vspace{-1.5\baselineskip}
    \caption{Histogram of output gradients (a) on value domain without Hadamard transform and (b) on frequency domain with Hadamard transform in ResNet-34 training on CIFAR-100.}
    \label{fig:lognormal}
    \vspace{-\baselineskip}
\end{figure}

\section{Alternative Optimization Technique: Hadamard Quantization}
In Section \ref{sec:LBP_limit}, we demonstrate that LBP-WHT has limitations in optimizing the \(g_x\) path. Therefore, alternative optimization techniques for \(g_x\) are necessary. Through extensive evaluations, we identify HQ as a promising candidate. In this section, we first explain how to apply HQ to BP and evaluate its sensitivity to \(g_x\) and \(g_w\).

Recent studies have shown that HT significantly reduces quantization error. For instance, when producing an output \(c = a \cdot b^T\) from arbitrary vectors \(a\) and \(b \in \mathbb{R}^n\), it can be modified as \(c = (a \cdot H_n) \cdot (H_n^T \cdot b^T)\) because the Hadamard matrix is orthogonal. In this equation, we first map the vectors to the frequency space, then produce an output via the inner product in frequency space, which remains consistent when using the full-rank Walsh-Hadamard matrix. However, when quantization is applied jointly, a surprising trend emerges: the output from \(Q(a \cdot H_n) \cdot Q(H_n^T \cdot b^T)\) shows a lower error than that from \(Q(a) \cdot Q(b)\), where \(Q(\cdot)\) denotes the quantization operator. Due to this property, many prior works have proposed HQ for optimizing inference. While these methods show promising results, to the best of our knowledge, the benefits of HQ have not been explored for BP. In the following section, we design the HQ pipeline for BP.

\subsection{Analysis on Hadamard Quantization for Backpropagation } \label{sec:HQ_limit}

\begin{wraptable}[11]{r}{0.45\textwidth}
    \centering
    \vspace{-\baselineskip}
    \scalebox{1}{
    \begin{tabular}{@{}ccc@{}}
    \toprule
    $g_i$ & $g_w$ & Accuracy (\%) \\ \midrule
   
    FP                 & FP            & 76.46                \\
    FP                  & 4-bit             & 72.09                \\
    FP              & 4-bit + HT         & 72.11                \\
    4-bit                  & FP            & 73.40               \\
    4-bit + HT              & FP            & 76.08                \\ \bottomrule
    \end{tabular}
    }
    \caption{Comparison of quantization effects on $g_w$ and $g_i$ in ResNet-50 for CIFAR-100.}
    \label{tab:quant effect}
\end{wraptable}

To apply HQ for BP, we introduce HT to Equation \ref{equ:bwd}. In the \(g_w\) path, HT in LBP-WHT can be used without modification, adjusting the \(g_w\) computation to \(\hat{g}_w = \frac{1}{B}(\bar{g}_y^T \cdot H_L^T) \cdot (H_L \cdot \bar{x}_L)\). Note that we apply the transformation to $L$ dimension when \(L\) is large enough, or apply \(H_B\) to the batch dimension instead. However, for \(g_x\), we apply the transformation to the \(O\) dimension to offset the Hadamard matrix during the inner product. Thus, the activation with HQ, \(\hat{g}_x\), is calculated via \(\hat{g}_x = Q(g_y \cdot H_O) \cdot Q(H_O^T \cdot w)\). Compared to native quantization, where \(\bar{g}_x = Q(g_y) \cdot Q(w)\), HQ shows lower quantization error, as presented in Figure \ref{fig:lognormal}. The mapped values in the frequency domain follow a smooth bell-shaped curve without outliers, allowing fine-grained, lower-error quantization.

To validate the benefit of HQ for BP, we conducted an ablation study analyzing the quality degradation induced by (Hadamard) quantization on the backpropagation of activation and weight. The results are presented in Table \ref{tab:quant effect}. When applying 4-bit stochastic min-max quantization to \(g_w\) or \(g_x\), notable accuracy degradation is observed. However, when applying Hadamard quantization, the quantization error is significantly reduced, resulting in a marked improvement in the trained model quality compared to naive quantization. Note that in Hadamard quantization, the same quantization operator is used after the transformation. While the transform introduces the mapping overhead, it provides strong benefits in terms of reduced quantization error.

Nonetheless, an important observation is that the damage to the weight gradient path \(g_w\) is not recoverable, even with Hadamard quantization. In the case of activations, multiple gradients are averaged via the batch dimension, allowing low-precision errors to be amortized. However, for weights, their gradients are directly accumulated to update the weights, so insufficient precision can steer the convergence trajectory and induce instability, resulting in significant quality degradation. Therefore, extensive quantization is not suitable for weight optimization.

\section{HLQ: Hadamard Low-rank Quantization}
In Sections \ref{sec:LBP_limit} and \ref{sec:HQ_limit}, our observations indicate that \(g_w\) can endure HLA but is highly sensitive to HQ. Conversely, \(g_x\) is robust to extensive HQ but suffers significantly from HLA. Therefore, we must apply optimization techniques selectively, considering the sensitivity of each computation path in BP.

In this study, we design a novel optimization pipeline for the BP of linear (or convolutional) layers, called Hadamard Low-rank Quantization (HLQ). Figure \ref{fig:HLQ} (c) visualizes the overview of the HLQ pipeline. In HLQ, we apply HLA and int8 acceleration for \(g_w\). Although \(g_w\) is sensitive to quantization, the Hadamard transform eliminates outliers in the gradient, enabling acceleration via int8 arithmetic with negligible quality loss. Moreover, a significant performance boost is expected because the low-rank projection itself induces negligible cost via FWHT but reduces FLOPs notably.

Conversely, we apply int4 HQ for \(g_x\) without low-rank approximation. HLQ preserves the full rank of the \(g_x\) path and batch-wise information, while providing substantial acceleration via int4 arithmetics because advanced NVIDIA GPUs support int4 matrix multiplication on TensorCore, offering up to 3x faster speed than int8 operators~\cite{wu2023understanding}. By leveraging the strengths of each method, we achieve maximal reduction of BP cost with negligible quality degradation. As demonstrated in the experiment section, our approach achieves comparable accuracy to the conventional training pipeline while providing up to 2.5 times faster computation and 78.50 \% lower memory footprint.

\subsection{Implementation Details}
Until now, we have explained the theoretical background of HLQ. However, to realize the benefits of HLQ in practice and maximize its advantages, various additional considerations have been applied. We will explain these detailed ideas to help readers understand.

First, like existing studies \cite{sun2020ultra, yang2024efficient}, we utilized block-diagonal Hadamard Transform (HT). Using the $H_4$ matrix as a basic building block, we constructed a 16 x 16 Hadamard matrix and then applied HT to the $L$ or $B$ dimensions for $g_w$ and to the $O$ dimension for $g_x$. For $g_w$, low-rank approximation needs to be considered. Following the $L_1$-base selection method of LBP-WHT, we selected 8 basis out of 16 channels from $H_4$, resulting in a reduction of the $L$ dimension to half.

For quantization, unbiased quantization is necessary because with biased quantization, the training can face notable quality degradation \cite{chmiel2022accurate}. It is known that stochastic rounding is an unbiased estimator \cite{chen2020statistical}, so HLQ also uses a min-max stochastic quantization. However, to reduce the overhead of generating random numbers, we use a pseudo stochastic quantizer \cite{wang2020niti}, utilizing the lower 11 bits of floating-point data as pseudo-random numbers to determine whether to round FP numbers.

In HLQ, $g_w$ computation is performed via six stages: 1. $x$ is projected to low-rank frequency space $\mathbb{R}^{\frac{L}{2} \times I}$ via HLA on $L$ dimension. 2. The projected $x$ is quantized into int8 precision. 3. $g_o^T$ is projected to low-rank frequency space $\mathbb{R}^{O \times \frac{L}{2}}$ via HLA on $L$ dimension. 4. The projected $g_o^T$ is quantized into int8.  5. Perform int8 matrix multiplication. 6. Dequantize the output to fp32. We utilize the CUTLASS TensorCore-based GEMM kernel for int8 acceleration. Additionally, the FWH transform, quantization, and dequantization stages are implemented to be performed on the shared memory to maximize the throughput of computation. 

In addition, to minimize the memory footprint during training, we apply stages 1 and 2 during forward propagation and store the intermediate feature map as an int4 representation. This simple technique reduces the memory footprint for activations to $\frac{1}{8}$ (fp32 → int4), and since activations take up a large portion of the overall memory space during training, this approach is highly effective in reducing memory usage. We call this technique Activation Compression for Backward Propagation (ACBP), and we apply it by default.

Likewise, $g_x$ computation is performed via six stages: 1. $g_o$ is projected to frequency space via FWH on $O$ dimension. 2. The projected $g_o$ is quantized into int4 precision. 3. $w$ is projected to frequency space via FWH on $O$ dimension. 4. The projected $w$ is quantized into int4 precision. 5. Perform int4 matrix multiplication. 6. Dequantize the output to fp32. We use CUTLASS TensorCore-based GEMM kernel and shared-memory based implementation as well.

\section{Experiments} \label{Experiment}
In order to validate the superiority of the proposed scheme, we conducted extensive experiments for two scenarios: model training from scratch and fine-tuning for transfer learning. 

For the training from scratch case, we trained ResNet-18, ResNet-34, and ResNet-50 as representative CNNs, and EfficientFormer-L1 and L3 as representative ViTs. These networks were trained on CIFAR-10, CIFAR-100, and ImageNet-100 datasets. For the fine-tuning case, we transferred pretrained EfficientNetV2-s, EfficientNetV2-m, EfficientFormer-L1, and L7 models from the ImageNet dataset to CIFAR-10 and CIFAR-100 datasets.

To compare the benefits of HLQ during training from scratch, we prepared three baselines: naive int4 quantization without Hadamard transform, LBP-WHT~\cite{yang2024efficient} that relies on HLA, and LUQ~\cite{chmiel2022accurate} that is the state-of-the-art method for fully quantized training. For fine-tuning, we used the same baselines for CNNs, but for ViT models, we used LoRA~\cite{hu2021lora} and LBP-WHT.

Our experiments were implemented using PyTorch 2.0 with custom CUDA kernels and conducted on a system with 8 $\times$ RTX 3090 GPUs. For full training, all models were trained for 200 epochs and updated with a batch size of 128. CNN models were trained with SGD with momentum. The learning rate was initially set to 0.1 and decayed by 1/10 at 60, 120, and 160 epochs. For the ViT models, they were trained with the AdamW optimizer \cite{loshchilov2017decoupled} and utilized the cosine annealing scheduler \cite{loshchilov2016sgdr}. During full training, all quantization operators use int8, including the baseline, at the initial 25 epochs to ensure a stable start of training. After the initial phase, the precision is modulated to the target bit-width, and training continues. For fine-tuning, the models are updated for 50 epochs using the AdamW optimizer with a learning rate of 0.001. Similar to the ViT case in full training, we utilize the cosine annealing scheduler.

\subsection{Comparison to the Trained Model Quality with Optimization}

% 표: accuracy 표
\begin{table}[]
\centering
\small
\begin{tabular}{@{}cclccccc@{}}
\toprule
\textbf{Task}               & \textbf{Model}                       & \multicolumn{1}{c}{\textbf{Dataset}} & \textbf{FP}                      & \textbf{int4} & \textbf{LUQ}                     & \textbf{LBP-WHT}                 & \textbf{HLQ}                     \\ \midrule
\multicolumn{8}{c}{Full Training}                                                                                                                                                                                                                                     \\ \midrule
                            &                                      & CIFAR10                              & 95.23                            & 93.1          & 94.73                            & 93.03                            & 94.58                            \\
                            &                                      & CIFAR100                             & 75.66                            & 75.22         & 75.63                            & 72.26                            & 75.69                            \\
                            & \multirow{-3}{*}{ResNet-18}          & IMG100                               & 87.19                            & NaN           & 87.39                            & 79.7                             & 85.2                             \\
                            &                                      & CIFAR10                              & 95.23                            & 93.57         & 94.74                            & 93.59                            & 95.12                            \\
                            &                                      & CIFAR100                             & 76.75                            & 72.87         & 76.26                            & 75.36                            & {\ul\textbf{76.35}}                            \\
                            & \multirow{-3}{*}{ResNet-34}          & IMG100                               & 88.64                            & NaN           & 88.51                            & 81.12                            & 88.32                            \\
                            &                                      & CIFAR10                              & 94.98                            & 90.65         & 93.62                            & 92.12                            & 94.11                            \\
                            &                                      & CIFAR100                             & 76.46                            & 61.28         & NaN                              & 69.24                            & 76.02                            \\
\multirow{-9}{*}{CNN}       & \multirow{-3}{*}{ResNet-50}          & IMG100                               & 89.28                            & NaN           & NaN                              & 82.28                            & 87.49                            \\ \midrule
                            &                                      & CIFAR10                              & 95.03                            & 92.9          & 94.1                             & 91.07                            & 94.79                            \\
                            &                                      & CIFAR100                             & 76.65                            & 64.15         & 75.79                            & 73.69                            & 76.25                   \\
                            & \multirow{-3}{*}{EfficientFormer-L1} & IMG100                               & 84.67                            & 81.3          & 82.46                            & 77.52                            & 82.57                            \\
                            &                                      & CIFAR10                              & 95.18                            & 93.8          & 94.63                            & 91.18                            & 94.37                            \\
                            &                                      & CIFAR100                             & 77.26                            & 66.93         & 76.19                            & 63.04                            & 76.94                            \\
\multirow{-6}{*}{ViT}       & \multirow{-3}{*}{EfficientFormer-L3} & IMG100                               & 88.13                            & 83.95         & 87.71                            & 78.6                             & {\ul\textbf{88.39}}                            \\ \midrule
\multicolumn{8}{c}{Fine Tuning}                                                                                                                                                                                                                                       \\ \midrule
                            &                                      & CIFAR10                              & 97.61                            & 95.95         & NaN                              & 94.09                            & 96.55                            \\
                            & \multirow{-2}{*}{EfficientNetV2-s}   & CIFAR100                             & 87.36                            & 84.89         & NaN                              & 87.1                             & {\ul\textbf{87.62}}                            \\
                            &                                      & CIFAR10                              & 98.01                            & 96.71         & NaN                              & 95.85                            & 97.07                            \\
\multirow{-4}{*}{CNN} & \multirow{-2}{*}{EfficientNetV2-m}   & CIFAR100                             & 84.89                            & 78.64         & NaN                              & 55.08                            & 82.37                            \\ \midrule
                            &                                      & CIFAR10                              & 95.23                            & \multicolumn{2}{c}{94.38 (LoRA)}                 & 94.6                             & 94.1                                \\
                            & \multirow{-2}{*}{EfficientFormer-L1} & CIFAR100                             & 79.28                            & \multicolumn{2}{c}{76.92 (LoRA)}                 & 78.27                            & {\ul\textbf{78.44}}                               \\
                            &                                      & CIFAR10                              & 97.61                            & \multicolumn{2}{c}{97.1 (LoRA)}                  & 97.22                            & 96.91                              \\
\multirow{-4}{*}{ViT}       & \multirow{-2}{*}{EfficientFormer-L7} & CIFAR100                             & 86.94                            & \multicolumn{2}{c}{85.09 (LoRA)}                 & 85.1                             & 85.83                              \\ \bottomrule
\end{tabular}
\caption{Quality degradation comprison with BP optimization techniques. The numbers in table represent the top-1 accuracy (\%). In Fine-tuning, the accuracy with LoRA is presented, instead of naive int4 quantization and LUQ. }
\label{tab:performance} 
\end{table}

The upper part of Table \ref{tab:performance} compares HLQ with other baseline methods in a full training environment. As seen in the table, int4 and LBP-WHT exhibit some performance degradation. However, HLQ shows the smallest quality degradation. For example, in the ResNet-34 experiment on CIFAR-100, LBP-WHT shows a 1.36\% accuracy drop, while HLQ only shows a 0.4\% drop. Similarly, in the EfficientFormer-L3 experiment on CIFAR-100, LUQ and LBP-WHT shows a 1.07\% and significant 14.22\% drop, respectively, whereas HLQ shows a 0.32\% loss.

The lower part of the table highlights HLQ's superiority in fine-tuning as well. According to the table, on EfficientNetV2-s with CIFAR-100, LUQ fails to train, and LBP-WHT shows a 0.26\% accuracy loss, while HLQ actually performs 0.26\% better than FP. This trend is particularly noticeable in fine-tuning ViT models. For EfficientFormer-L1 on CIFAR-100, HLQ outperforms both LoRA and LBP-WHT by a large margin. Moreover, It is also noteworthy that LUQ, the state-of-the-art model in traditional gradient quantization, shows similar quality in some models, but as the model depth increases, the training becomes significantly unstable, often resulting in training failures. Compared to existing methods, HLQ not only provides the highest quality but also the greatest stability.

\begin{figure}
    \centering
    \includegraphics[width=\textwidth]{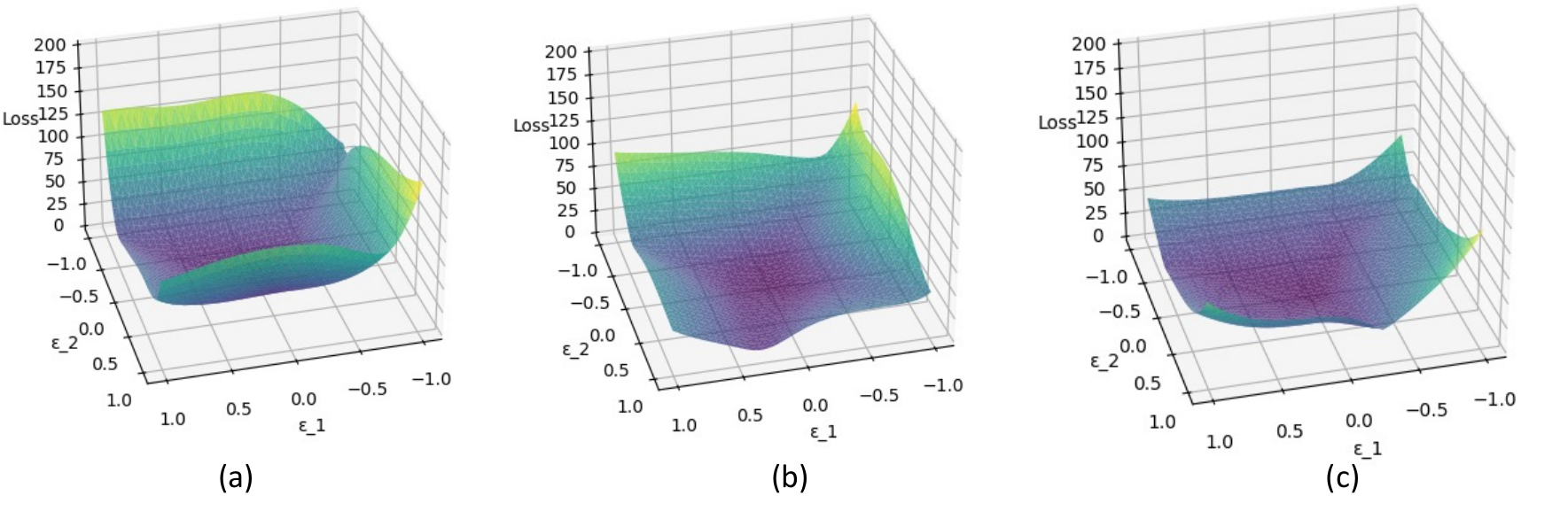}
    \vspace{-1.5\baselineskip}
    \caption{The loss landscape visualization of ResNet-34 on CIFAR-100 with (a) naive int4 quantization, (b) int4 with HT, and (c) HLQ.}
    \label{fig:loss}
    \vspace{-\baselineskip}
\end{figure}

For a more comprehensive analysis of the effect of HQ and the benefit of HLQ, we visualize the loss landscape~\cite{li2018visualizing} of the trained model, presented in Figure \ref{fig:loss}. Compared to figure (a), figure (b) exhibits a smoother surface with lower values, which explains the benefit of HQ for BP optimization. Additionally, among these, figure (c) demonstrates the lowest loss surface, visualizing the benefit of HLA for the $g_w$ path instead of applying int4 quantization.

\subsection{Computation Cost Reduction Analysis}

\begin{table}[]
\small
\centering
\begin{tabular}{@{}cr@{}}
    \toprule
    \multicolumn{1}{c}{\textbf{Name}} & \multicolumn{1}{c}{\textbf{FLOPs}} \\ \midrule
    Vanilla BP                        & $4LIO$                            \\ \midrule
    $g_x$                      & \makecell[r]{$2LOlogn + 2IOlogn$ \\ $+2LO + 2IO$}                          \\ \midrule          
    $g_w$                      & \makecell[r]{$2LIlogn + 2LOlogn$\\$+2I(L*\frac{r}{n}) + 2O(L*\frac{r}{n})$}                            \\ \midrule
    Dequant                           & $2IO + 2LI$                             \\ \bottomrule
\end{tabular}
\quad
\begin{tabular}{@{}cccccc@{}}
\toprule
\textbf{Model}      & \textbf{FP} & \textbf{LBP-WHT}  & \textbf{HLQ}\\ \midrule
ResNet-18           & 2328.78     & 468.98           & 219.18       \\
ResNet-34           & 4802.76     & 863.79           & 404.11       \\
ResNet-50           & 5531.17     & 1020.3           & 501.83       \\ \midrule
EfficientForemer-L1 & 5276.55     & 524.19           & 334.52      \\
EfficientForemer-L3 & 15951.69    & 1301.95          & 801.67       \\ \midrule
EfficientNetV2-s    & 11689.09    & 1169.51          & 795.37       \\
EfficientNetV2-m    & 21982.80    & 2515.34          & 1532.57      \\ \bottomrule
\end{tabular}
\caption{(Left) The additional FLOPs induced by optimization path and (Right) the estimation of computation cost. The numbers in the table represents the $10^9$ binary operations (GBoPs).}
\label{tab:BOPS} 
\end{table}

\begin{table}[]
\small
\centering
\begin{tabular}{@{}cccc|cccc@{}}
\toprule
\multicolumn{4}{c|}{\textbf{EfficientFormer-L1}}                                      & \multicolumn{4}{c}{\textbf{ResNet-34}}                                                  \\ \midrule
\textbf{(L, O, I)} & \textbf{FP} & \textbf{HLQ} & \textbf{Speed Up}      & \textbf{(L, O, I)} & \textbf{FP} & \textbf{HLQ} & \textbf{Speed Up}      \\ \midrule
(196, 224,   896)  & 166                  & 87                    &$\times$1.9               & (1024, 64, 576)    & 194                  & 95                    &$\times$2.0                 \\ \cmidrule(l){5-8} 
(196, 896,   224)  & 185                  & 90                    &{\ul\textbf{$\times$2.1}}                 & (256, 128, 152)    & 156                  & 76                    &$\times$2.1                 \\ \cmidrule(r){1-4}
(196, 224,   864)  & 182                  & 91                    &$\times$2.0                 & (256, 128, 64)     & 111                   & 50                    &$\times$2.2                \\ \cmidrule(l){5-8} 
(196, 864,   224)  & 170                  & 84                    &$\times$2.0                 & (64, 256, 1152)    & 171                  & 86                    &$\times$2.0                 \\ \cmidrule(r){1-4}
(784, 96, 432)     & 181                  & 88                    &{\ul\textbf{$\times$2.1}}                 & (64, 256, 128)     & 113                   & 50                    &{\ul\textbf{$\times$2.3}}  \\ \cmidrule(l){5-8} 
(784, 96, 384)     & 179                  & 91                    &$\times$2.0                 & (16, 512, 2304)    & 252                  & 101                    &{\ul\textbf{$\times$2.5}}                 \\ \bottomrule
\end{tabular}
\caption{Latency profiling results measured on real GPU with varying layer dimensions. The numbers in table represent the latency ($\mu$s).}
\label{tab:prof}
\end{table}

The additional transformation, reshaping, and quantization/dequantization processes introduce some overhead to HLQ compared to the vanila BP path. However, by leveraging low-precision arithmetic, we can achieve practical performance benefits. In this section, we present three tables to explain the computational benefits of our approach in detail.

Compared to vanilla BP, the added overhead of HLQ can be negligible, as shown in Table \ref{tab:BOPS} (left), especially when $\log n$ is sufficiently small relative to other dimensions. In our work, we use $n = 16$, making this condition valid. For example, in the second-to-last layer of EfficientFormer-L1, vanilla BP requires 137.3 MFlops, whereas our method only requires 11.5 MFlops. By incurring this 7\% overhead, we can utilize efficient low-precision arithmetic.

While Table \ref{tab:BOPS} (left) presents the additional overhead, the overall cost including integer arithmetic is not evaluated. To compare the end-to-end computation cost, we evaluate the training cost of various models using bit-operations (BoPS)~\cite{uniq, shin2023nipq} as a metric to measure the computation cost for both floating-point and integer operations. Table \ref{tab:BOPS} (right) 
shows the results, and HLQ only takes less than 9\% of the BOPS compared to vanilla BP, highlighting the efficiency of the proposed algorithm.

In addition to the numerical estimation, we performed additional experiments to estimate the practical performance improvement on a real GPU. For a more practical observation, we implemented a specialized kernel and profiled the latency on the real GPU. Based on the profiled numbers, our method can achieve up to 2.5 times acceleration on a real GPU compared to the vanilla path. Since matrix multiplication occupies the majority of the execution cycle, accelerating it through low-precision arithmetic provides a satisfactory improvement.

\begin{table}[]
\small
\centering
\begin{tabular}{@{}cccccc@{}}
\toprule
\textbf{Model}      & \textbf{FP} & \textbf{LBP-WHT} & \textbf{HLQ  (no ACBP)} & \textbf{HLQ} & \textbf{Memory Reduction} \\ \midrule
ResNet-18           & 2055.04     & 1794.54          & 1248.23                & 449.22       & 78.14\%                    \\
ResNet-34           & 3872.23     & 3392.89          & 2346.78                 & 832.54       & {\ul\textbf{78.50\%}}                    \\
ResNet-50           & 4309.22     & 3892.42          & 2756.23                 & 1164.96      & 72.97\%                    \\ \midrule
EfficientFormer-L1 & 2962.94     & 2265.08          & 2006.83                 & 1112.28      & 62.46\%                    \\
EfficientFormer-L3 & 6966.65     & 4707.92          & 4692.24                 & 2479.07      & 64.42\%                    \\
EfficientFormer-L7 & 16455.02    & 9203.84          & 10973.14                 & 5445.30      & {\ul\textbf{66.91\%}}                    \\ \midrule
EfficientNetV2-s    & 6563.11     & 5381.21          & 4385.94                 & 2496.88      & 61.96\%                    \\
EfficientNetV2-m    & 12844.28    & 11113.47         & 8525.41                 & 4445.02      & 65.39\%                    \\ \bottomrule
\end{tabular}
\caption{Memory footprint reduction of training at CIFAR-100 dataset. $B = 64$, and the numbers in the table represents the memory size (MB).}
\label{tab:Memory reduction} 
\end{table}

\subsection{Memory Reduction Analysis}
Another important benefit of HLQ is its high memory efficiency. By compressing the data representation via HQ and HLA, we significantly reduce the memory footprint. Compared to LBP-WHT, when we apply HLQ without ACBP, memory consumption is comparable. However, the simple idea of ACBP reduces the memory space for intermediate activation to 1/8 (from fp32 to int4), significantly lowering memory usage. Table \ref{tab:Memory reduction} presents the memory reduction achieved by HLQ. HLQ reduces the memory usage of CNN models by up to 78.5\% compared to fp32, and for ViT models, up to 66\%. Compared to LBP-WHT, HLQ enables training only using 59.2\% or less memory compared to the vanilla BP (or up to 75.5\% lower memory usage than LBP-WHT), showing its superiority.

\subsection{Limitation}
While our study effectively demonstrated the benefits of HLQ on computer vision models, we did not experimentally validate its effectiveness on LLMs. However, our framework suggests that HLQ could be applicable to LLMs. Therefore, we intend to investigate the extension of HLQ to LLMs.

\section{Conclusion}
In this study, we propose a novel efficient training scheme called Hadamard Low-rank Quantization (HLQ). We focus on optimizing the backward pass operations by evaluating the potential of Hadamard quantization and Hadamard low-rank approximation for the gradient computation paths of activations and weights. These two paths have different sensitivities to these optimization schemes, so we design HLQ judiciously to minimize the quality degradation of the trained model while maximizing acceleration and storage reduction benefits. Extensive analysis shows the superiority of the proposed scheme, achieving up to 2.5x acceleration and 78.5\% memory reduction.

\bibliographystyle{unsrt}
\bibliography{hlq_arxiv24.bbl}
\end{document}